\documentclass[conference]{IEEEtran}
\IEEEoverridecommandlockouts
\usepackage{cite}
\usepackage{mathtools}
\usepackage{amsmath,amssymb,amsfonts}
\usepackage{graphicx}
\usepackage{textcomp}
\usepackage{xcolor}
\usepackage{amsmath}
\usepackage{multirow}
\usepackage{multicol}
\usepackage{caption}
\usepackage{array}
\usepackage{subcaption}
\usepackage{balance}
\usepackage{wrapfig}
\usepackage{hyperref}
\usepackage{multirow}
\usepackage{multicol}
\usepackage{color}
\usepackage{parskip}
\usepackage{colortbl}
\usepackage{booktabs}
\usepackage{flushend}
\usepackage{hyperref}
\usepackage{todonotes}
\usepackage{comment}
\usepackage{cite}

\usepackage{algpseudocode}
\usepackage{algorithm}
\newcolumntype{L}{>{\centering\arraybackslash}m{3cm}}

\makeatletter
\newcommand{\linebreakand}{%
  \end{@IEEEauthorhalign}
  \hfill\mbox{}\par
  \mbox{}\hfill\begin{@IEEEauthorhalign}
}
\makeatother

\def\BibTeX{{\rm B\kern-.05em{\sc i\kern-.025em b}\kern-.08em
    T\kern-.1667em\lower.7ex\hbox{E}\kern-.125emX}}
\begin{document}

\title{Topic-Centric Unsupervised Multi-Document Summarization of Scientific and News Articles\\

\thanks{This effort was sponsored in whole or in part by the Air Force Research Laboratory, USAF, under Memorandum of
Understanding/Partnership Intermediary Agreement No FA8650-18-3-9325. The U.S. Government is
authorized to reproduce and distribute reprints for Governmental purposes notwithstanding any copyright
notation thereon.}
}

\author{\IEEEauthorblockN{Amanuel Alambo}
\IEEEauthorblockA{\textit{Computer Science and Engineering} \\
\textit{Wright State University}\\
Dayton, OH \\
alambo.2@wright.edu}
\and
\IEEEauthorblockN{Cori Lohstroh}
\IEEEauthorblockA{\textit{ONEIL Center} \\
\textit{Wright State University}\\
Dayton, OH \\
lohstroh.2@wright.edu}
\and
\IEEEauthorblockN{Erik Madaus}
\IEEEauthorblockA{\textit{ONEIL Center} \\
\textit{Wright State University}\\
Dayton, OH \\
madaus.2@wright.edu}
\and
\IEEEauthorblockN{Swati Padhee}
\IEEEauthorblockA{\textit{Computer Science and Engineering} \\
\textit{Wright State University}\\
Dayton, OH \\
padhee.2@wright.edu}
\linebreakand %

\IEEEauthorblockN{Brandy Foster}
\IEEEauthorblockA{\textit{ONEIL Center} \\
\textit{Wright State University}\\
Dayton, OH \\
brandy.foster@wright.edu}
\and
\IEEEauthorblockN{Tanvi Banerjee}
\IEEEauthorblockA{\textit{Computer Science and Engineering} \\
\textit{Wright State University}\\
Dayton, OH \\
tanvi.banerjee@wright.edu}
\and
\IEEEauthorblockN{Krishnaprasad Thirunarayan}
\IEEEauthorblockA{\textit{Computer Science and Engineering} \\
\textit{Wright State University}\\
Dayton, OH \\
t.k.prasad@wright.edu}
\linebreakand %
\IEEEauthorblockN{Michael Raymer}
\IEEEauthorblockA{\textit{Computer Science and Engineering} \\
\textit{Wright State University}\\
Dayton, OH \\
michael.raymer@wright.edu}
}

\maketitle

\begin{abstract}
Recent advances in natural language processing have enabled automation of a wide range of tasks, including machine translation, named entity recognition, and sentiment analysis. Automated summarization of documents, or groups of documents, however, has remained elusive, with many efforts limited to extraction of keywords, key phrases, or key sentences. Accurate abstractive summarization has yet to be achieved due to the inherent difficulty of the problem, and limited availability of training data. In this paper, we propose a topic-centric unsupervised multi-document summarization framework to generate extractive and abstractive summaries for groups of scientific articles across 20 Fields of Study (FoS) in Microsoft Academic Graph (MAG) and news articles from DUC-2004 Task 2. The proposed algorithm generates an abstractive summary by developing salient language unit selection and text generation techniques. Our approach matches the state-of-the-art when evaluated on automated extractive evaluation metrics and performs better for abstractive summarization on five human evaluation metrics (entailment, coherence, conciseness, readability, and grammar). We achieve a kappa score of 0.68 between two co-author linguists who evaluated our results. We plan to publicly share MAG-20, a human-validated gold standard dataset of topic-clustered research articles and their summaries to promote research in abstractive summarization.
\end{abstract}


\section{Introduction}
With the large number of articles published in the research and media community, there is an increasing demand to produce summaries that are coherent, concise, informative, and grammatical. Summarization comes in two forms: \textit{extractive} and \textit{abstractive}. Extractive summarization \cite{zhou2020joint, liu2019fine, narayan2018ranking} is focused on extracting significant sentences \cite{thirunarayan2007selecting} from source documents and has been well studied. Abstractive summarization aims at ways of fusing or paraphrasing sentences in source documents to form abstractive sentences. Due to the challenges of capturing abstractive concepts shared among sentences across source documents, and synthesizing an informative summary, there has been limited progress in abstractive summarization. While there are recent advances in unsupervised multi-document abstractive summarization \cite{chu2019meansum}, \cite{nayeem2018abstractive}, \cite{banerjee2015multi}, they are usually limited to forming summaries by copying words in source documents and re-arranging the words to form new sentences. These approaches identify salient phrases from sentences in source documents and fuse them to form abstractive summaries. Thus, they do not perform abstraction of sentences.

Our framework consists of two phases: an \emph{extractive phase} and an \emph{abstractive phase}. In the extractive phase, we follow a three-fold approach. First, we identify the core article and peripheral articles for each set of related articles. Second, we instantiate clusters using the language units of a core article and perform centroid based clustering to place language units from peripheral articles into the clusters initialized by the language units in a core article. Third, we fuse language units in a cluster using an enhanced Multi-Sentence Compression (MSC) \cite{filippova2010multi}, \cite{zhao2019unsupervised} technique with a novel algorithm to maximize topical coverage and relevance of a path to the language units in the cluster. In the abstractive summarization phase 1) we employ text generation to generate abstractive language units (ALUs); and 2) we use MSC to fuse the generated ALUs into an abstractive summary. Unlike DUC-2004, where articles are topically clustered, scientific articles in MAG-20 \cite{sinha2015overview} do not come topically-grouped. Therefore, for MAG-20, we use topical hierarchical agglomerative clustering (HAC) to cluster articles. 

The key contributions of our study are 1) an ALU generation technique using GPT-2; 2) a novel MSC-based algorithm for selection of informative paths; and 3) a gold standard dataset of topical clusters of articles from MAG-20 and their multi-doc abstractive summaries. We use abstracts of articles from MAG for this study.

\section{Related Work}
Different techniques have been proposed for unsupervised abstractive summarization, including sequence to sequence models \cite{shi2018neural}, \cite{khatri2018abstractive}, neural models with and without attention \cite{chu2019meansum}, \cite{liu2018generating}, \cite{see2017get}, abstract meaning representation (AMR) \cite{liao2018abstract}, \cite{liu2018toward}, and centroid-based summarization \cite{nayeem2018abstractive}, \cite{banerjee2015multi}. Our approach extends the state of the art techniques in centroid-based summarization by employing language unit identification from articles and a novel text generation technique. 

Abstractive summarization has received significant attention due to progress in deep representation learning \cite{vaswani2017attention}, \cite{devlin2018bert}. \cite{chu2019meansum} propose MeanSum, which consists of an autoencoder and a summarization module to produce abstractive summaries. The abstractive summarization approach of \cite{banerjee2015multi} called ILPSumm includes identification of informative content and clustering of similar sentences from the documents to form summaries. \cite{nayeem2018abstractive} extended the technique proposed in \cite{banerjee2015multi} by introducing a paraphrastic fusion model, called ParaFuse, based on context-sensitive substitution of target words. While lexical substitution enables the generation of novel words, it is limited when it comes to capturing the context in the source document. \cite{see2017get} propose a Pointer Generator Network to summarize news articles from the CNN/Dailymail dataset.  

Further, \cite{krishna2018generating} propose a framework that takes an article and a topic and generates a summary specific to the topic. However, their work is supervised and relies on the availability of human-generated training corpus to train their model.

\section{Data Collection}
We work with two datasets to better understand and evaluate our proposed approach: 1) the DUC-2004 benchmark dataset; and 2) scientific articles from MAG.

We queried MAG for the 100 most-cited abstracts for each of the 20 FoS published in 2016 - 2020. The 20 FoS we used are: Artificial Intelligence, Artificial Neural Network,  Big Data, Case-Based Reasoning, Cybernetics, Cyberwarfare, Data Mining, Data Science, Decision Support System, Electronic Warfare, Expert System, Human-Machine Interaction, Intelligent Agent, Knowledge-Based Systems, Machine Learning, Multi-Agent System, Prediction Algorithms, Predictive Analytics, Predictive Modeling, and Sensor Fusion.

\section{Proposed Method}

\subsection{Extractive Phase}

Fig. 1 shows the sequence of steps we devised for MAG-20 and DUC-2004 extractive summarization. The difference in the extractive phase of these tasks is MAG-20 has topic modeling and hierarchical agglomerative clustering in its pipeline.

\begin{figure}[htbp]
\centerline{\includegraphics[width=70mm]{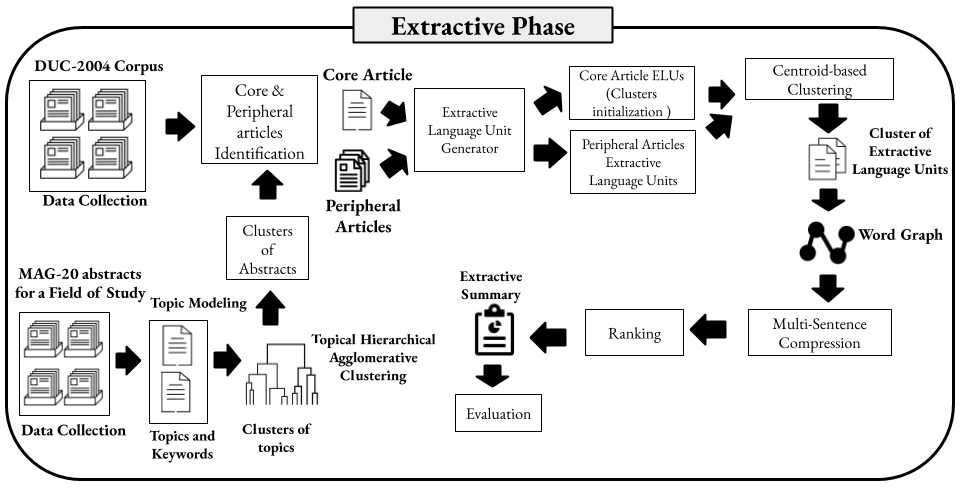}}

\caption{Extractive Summarization. The pipeline following \emph{Cluster of Abstracts} is for each cluster of topics.}

\label{fig}
\end{figure}

\subsubsection{Topic Modeling}

For MAG-20, we determine groups of topically related abstracts for each FoS. We first build LDA \cite{blei2003latent} topic models for an FoS using number of topics in the range of 2 to 92. We then determine the optimal number of topics from an ensemble of the LDA models that maximizes the coherence score. The topics, and thus, keywords generated using the LDA model that gives the highest coherence score, are used for Topical HAC.

\subsubsection{Topical Hierarchical Agglomerative Clustering}
Different topics generated using an LDA model can have semantically redundant keywords.  We thus cluster topics having high similarity among their keywords using HAC. We use SciBERT \cite{beltagy2019scibert} embeddings to represent each keyword in a topic. A topic is then represented as a concatenation of the representations of its keywords. 
Once each topic is represented, we conduct topical HAC. To determine the number of clusters for a collection of topics, we ran HAC for several clusters ranging from 2 to the total number of topics. We use Silhouette coefficient to determine the optimal number of clusters. \\

We introduce a topical similarity metric (Equation-1) for measuring the similarity between a pair of topics. Each keyword in a topic is compared with all the keywords in another topic, and the sum of highest similarity scores is preserved.

\begin{equation}
    \begin{aligned}
        sim(\mbox{Topic-i, Topic-j}) = \sum_{i \in \mbox{Topic-i}} maxcos(i, \mbox{Topic-j}) \\
    \end{aligned}
\end{equation}

where\\
$maxcos(i, \mbox{Topic-j}) = \mbox{maximum of cosine similarities between} \\
\mbox{term } i \mbox{ and terms in Topic-j}$

\begin{table}[ht]
\begin{center}
\begin{tabular}{|c|c|c|}
\hline
 \textbf{Cluster} & \textbf{Topic IDs} \\ 
 \hline
 0 & 0, 1, 6, 7, 8, 9, 14, 21  \\  
 1 & 2,  3,  4,  5, 13, 17, 20 \\
 2 & 10, 11, 12, 15, 16, 18, 19 \\
 \hline
\end{tabular}
\end{center}
\caption{Topics and their cluster membership}
\end{table}

A MAG-20 abstract is assigned to a topic that is the most dominant among all possible topics the abstract addresses. Table I shows the three clusters and their constituent topic IDs. Table II shows topical distribution among abstracts for a selected field of study. It can be seen two abstracts have the same dominant topic. These abstracts form a set of documents on which we perform multi-document summarization. \\

\begin{table}[ht]
\begin{center}
\scalebox{0.65}{
\vline
\begin{tabular}{p{1.0cm}|p{1.0cm}|p{2.5cm}|p{5.0cm}|}

\hline
 \textbf{Abstract ID} & \textbf{Dominant Topic} & \textbf{Dominant Topic Contribution(\%)} & \textbf{Topic 
Keywords} \\

 \hline
 6 & 19 & 0.87 & inspire, state, device, accelerator, small, size, high, power, ved, advantage \\
 
 \hline
 4 & 19 & 0.59 & inspire, state, device, accelerator, small, size, high, power, ved, advantage \\
 
 \hline
\end{tabular}
}
\end{center}
\caption{Abstracts and Dominant Topic.}
\end{table}

\subsubsection{Core and Peripheral Articles Identification}
We identify the core article from a cluster of articles based on how similar an article is to other articles. Equation-2 computes the Cross-Article Similarity Score of an article. An article with the highest cumulative similarity with other articles in a cluster is chosen as the core article. The rest of the articles in the cluster are peripheral articles.

\begin{equation}
    \begin{aligned}
        CAS_i = \frac{\sum_{{i,j} \in C} doc2vec\_sim(i, j)}{N}
\label{eq:cross_abstract_score}
    \end{aligned}
\end{equation}

\noindent where
$i \ne j\\
N \mbox{- Number of articles in the cluster}\\
C \mbox{- The cluster of articles}\\
doc2vec\_sim$ \mbox{- doc2vec-based cosine similarity}

\subsubsection{Centroid based Clustering}
After core and peripheral articles are identified, we generate extractive language units (ELUs) from the core and the peripheral articles. Recent studies in centroid based summarization utilized sentences in documents as standalone ELUs to initiate clusters and to quantify semantic relatedness \cite{nayeem2018abstractive}, \cite{banerjee2015multi}. However, this approach breaks the interdependence among sentences in a document and eventually leads to incoherent summaries. We address this issue by identifying the sentences that are interdependent using neural coreference resolution \cite{lee2017end} and preserving them as one ELU. Once the ELUs from the core article have instantiated clusters, the ELUs from the peripheral articles are placed into a cluster based on the cosine similarity between the embedding of an ELU from the peripheral article and the embeddings of the ELUs from the core article. An ELU embedding is constructed by concatenating the embeddings of the sentences using sent-BERT \cite{reimers2019sentence} and performing dimensionality reduction to 300 units using T-SNE. The purpose of dimensionality reduction is to have a uniform dimension among ELUs even when they contain different number of sentences so that cosine similarity can be computed. 

\subsubsection{Multi-Sentence Compression}
The number of clusters formed in the centroid-based clustering stage is the same as the number of ELUs in the core article. After clusters of ELUs are formed, we build word graphs \cite{boudin2013keyphrase} for each cluster. Fig. 2 shows a sample word graph constructed for a cluster consisting of the following ELUs:\\

\noindent $ELU_{1}$ = "Radars are required to limit emissions in adjacent bands, but traditional rectangular  pulses have high out-of-band emissions."\\
$ELU_2$ = "Millimeter wave radars are popularly used in last-mile radar based defense systems.”\\

\begin{figure}[htbp]
\centerline{\includegraphics[width=60mm]{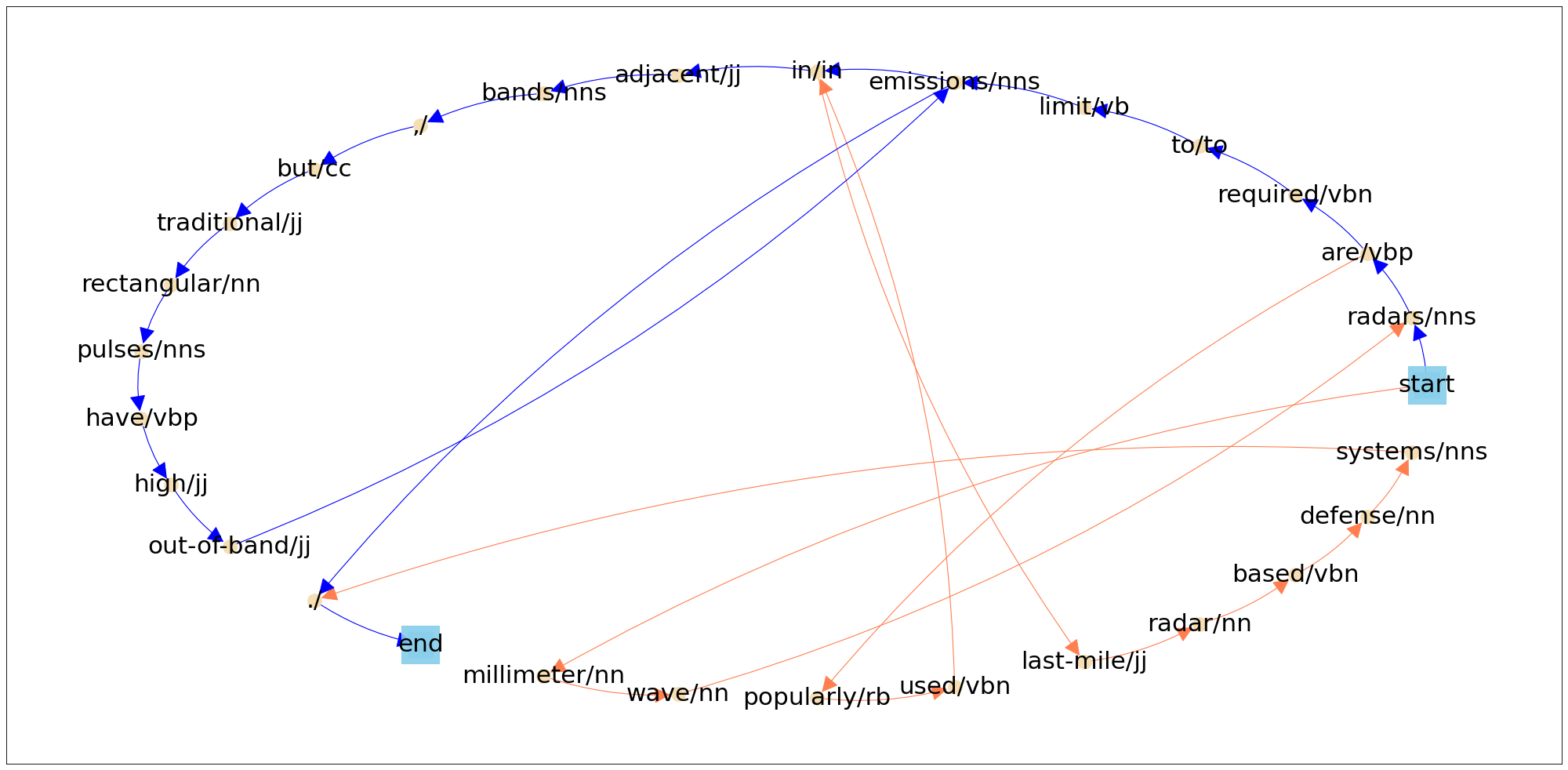}}

\caption{Word Graph for two ELUs using NetworkX. Tokens and PoS tags of the tokens are used for a node.}

\label{fig}
\end{figure}

We develop an algorithm for extracting paths based on topical coverage and relevance. A path is selected using an additional criterion that a candidate path should at least span two ELUs in the cluster. Next, we generate topically informative and relevant paths from the word graph while maintaining the 100-word summary limit. Topical coverage (Equation-3) measures how well a path covers the dominant topics discussed by the articles of the ELUs. Relevance (Equation-4) measures how relevant a path is to the ELUs. The cumulative score of a path (Equation-5) is determined by a weighted sum of topical coverage and relevance. We experimented with values of $\alpha$ in the range of 0 to 1.\\

\paragraph{Topical Coverage Formulation}

\begin{equation}
    \begin{aligned}
         Coverage(C_{path}, C_{topics}) 
         = \frac{1}{|C_{path}|} \sum_{i_{C_{path}} \in C_{path}} ^ {|C_{path}|} \frac{1}{|C_{topics}|} \\ * \sum_{K_c \in C_{topics}} ^ {|C_{topics}|} \emph{ maxcos}(i_{C_{path}}, K_c)
    \end{aligned}
\end{equation}

\noindent where,
$C_{path}$ \mbox{- Candidate path}\\
$C_{topics}$ \mbox{- Cluster of topics}\\

\noindent Topical coverage is measured with respect to the cluster of topics.\\

\paragraph{Path Relevance Formulation}
\begin{equation}
    \begin{aligned}
         Relevance(C_{path}, C_{ELU}) = \frac{\vec{v}(C_{path})\cdot\vec{v}(C_{ELU})}{|\vec{v}(C_{path})|\cdot|\vec{v}(C_{ELU})|}
    \end{aligned}
\end{equation}

\noindent where,
$C_{path}$ \mbox{- Candidate Path}\\
$C_{ELU}$ \mbox{- Cluster of ELUs}\\
$\vec{v}(C_{path})$ \mbox{- Vectorial Representation of Candidate Path}\\
$\vec{v}(C_{ELU})$ \mbox{- Vectorial Representation of Cluster of ELUs}\\

Path relevance is measured with respect to the ELUs. 

\paragraph{Cumulative Score}
\begin{equation}
    \begin{aligned}
        \mbox{Score}(C_{path}) = 
        \alpha \cdot \mbox{Coverage}(C_{path}, C_{topics}) \\
        + (\mbox{1} - \alpha) \cdot \mbox{Relevance}(C_{path}, C_{ELU})
    \end{aligned}
\end{equation}

A path is selected from the word graph 1) if the path is longer than the average minimum length of a sentence in an FoS or DUC-2004 topic and smaller than the average maximum length of a sentence; 2) if the combined topical coverage and relevance for the path meets or exceeds a threshold $\tau$ of 0.5. If a path picked from the word graph is semantically similar to an already selected path by an order of threshold $\delta$ of 0.8 or more, we compare the combined topical coverage and relevance of the two paths and keep the one with a higher score and remove the other. The selection of 0.8 is based on empirical observations.

\subsection{Abstractive Phase}

Fig. 3 shows the steps we followed for abstractive summarization. The difference in the abstractive phase of MAG-20 and DUC-2004 is DUC-2004 has headline generation component.

\begin{figure}[htbp]
\centerline{\includegraphics[width=70mm]{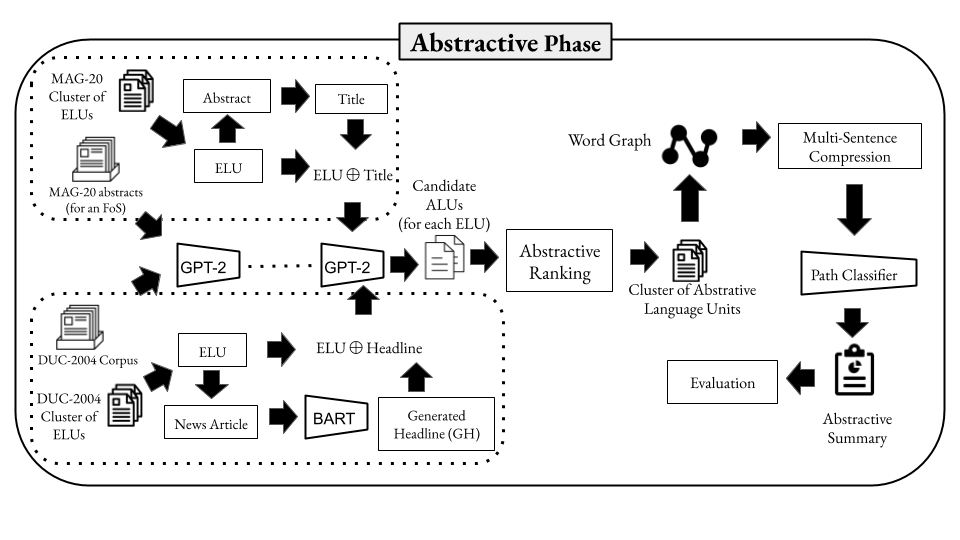}}

\caption{Abstractive Summarization Pipeline.}

\label{fig}
\end{figure}
\subsubsection{Abstractive Language Unit (ALU) Generation}

We start our abstractive phase with a pragmatic assumption that the title/headline of an article is an abstraction of the individual extractive language units (ELUs) within the same article. We propose a method to generate an ALU for an ELU using the ELU and title/headline as prompts for generating text. Combining bidirectional encodings of the title/headline with an ELU enables generating abstractive text. For ELUs consisting of two or more sentences, we encode each sentence using sentence-BERT \cite{reimers2019sentence}, and then we concatenate these representations. Next, we perform dimensionality reduction using T-SNE to encode an ELU. For encoding a title/headline, we use sentence-BERT without dimensionality reduction. We fine-tune a GPT-2 model for an FoS (Fig. 4) and use the fine-tuned GPT-2 model to generate ALUs given a concatenation of the bidirectional encodings of the ELU and the title/headline of an article. We fine-tune a GPT-2 model such that it has 124M parameters and generates 10 candidate ALUs. While fine-tuning, we set the temperature to 0.7, number of generated samples to 10, top\_k random sampling to 2 to generate more ALUs and minimize redundancy \cite{radford2019language}. We train the GPT-2 for 10 epochs with a batch size of 10 and attain a loss of 2.16. We select an ALU that maximizes semantic similarity and minimizes syntactic similarity with the ELU used for generation. We use the normalized sum of ROUGE-1($R_1$) and ROUGE-2 ($R_2$) for syntactic similarity. We introduce an \emph{abstractiveness score} for an ALU, as shown in Equation-6.

We use BART \cite{lewis2019bart} for headline generation for each DUC-2004 article that is later used for ALU generation along with an ELU.

\begin{figure}[htbp]
\centerline{\includegraphics[width=50mm]{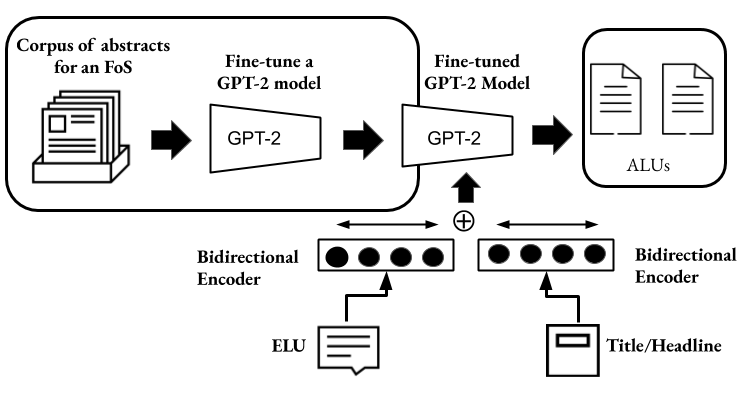}}

\caption{ALUs generation using GPT-2.}

\label{fig}
\end{figure}

\begin{equation}
    \begin{aligned}
        \mbox{Score}(\mbox{ALU, ELU}) = 
        \mbox{cossim}_{\mbox{xBERT}}(\mbox{ALU, ELU}) \\
        - \frac{[R_1(\mbox{ALU, ELU}) + R_2(\mbox{ALU, ELU}) }{[R_1(\mbox{ALU, ALU}) + R_2(\mbox{ALU, ALU}) }
    \end{aligned}
\end{equation}

where\\
\mbox{ALU} \mbox{- } \mbox{Abstractive Language Unit}\\
\mbox{ELU} \mbox{- } \mbox{Extractive Language Unit}\\
$\mbox{cossim}_{\mbox{xBERT}}$ \mbox{- Cosine similarity on x-dimension BERT} \\ 
\mbox{embeddings}

We select an ALU that gives the highest \emph{abstractiveness score} (Equation-6) from candidate ALUs. Table III shows a sample ELU and highest scoring ALU generated.

\begin{table}

\label{tab:example}
\centering

\begin{tabular}{| p{3.5cm} | p{2.5cm} |}
    \hline
    \textbf{ELU}  &  \textbf{ALU}\\
    \hline
    
    The ability to repair relationship and work together will be the key to a stable coalition.   &   It's a good time for a new political party that can bring stability and development.\\
    \hline

\end{tabular}

\caption{ALU generated using GPT-2.}

\end{table}

\subsubsection{Multi-Sentence Compression}

After generating ALUs for a cluster, we build a word graph and run our MSC algorithm as used in the extractive phase; i.e., the same ranking formulation and path selection algorithm is used for selecting informative paths from a word graph built, this time from a cluster of ALUs. Fig. 5 shows a cluster of ALUs and the generated fused paths that form the final abstractive summary.

\begin{figure}[htbp]
\centerline{\includegraphics[width=65mm]{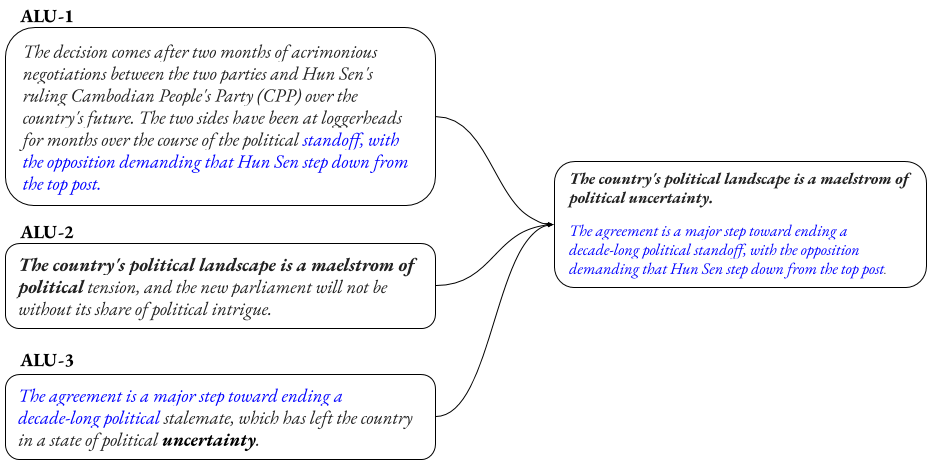}}

\caption{Candidate ALUs and compressed ALU paths.}

\label{fig}
\end{figure}

\section{Results and Discussion}

\subsection{Extractive Evaluation}
We use ROUGE metrics for evaluating extractive summaries taking the source articles as the reference summary. 

DUC-2004 and MAG-20 extractive evaluation results are shown in Tables IV and V, respectively. It can be seen that our proposed method performs comparably to the baseline approaches on ROUGE-1, ROUGE-2, and ROUGE-L metrics. 

\begin{table}
\label{tab:example}
\centering
\begin{tabular}{|c|c|c|c| }
    \hline
    \textbf{Model}  &  \textbf{R-1} & \textbf{R-2} & \textbf{R-L}\\
    \hline

    ILPSumm  &  39.24 & 11.99 & 9.34\\
    \hline
    
    ParaFuse  &  \textbf{40.07} & \textbf{12.04} & \textbf{11.28}\\
    \hline
    
    Our Proposed Method  &  39.58 & 11.36 & 9.83\\
    \hline
    
\end{tabular}

\caption{DUC-2004 Extractive Evaluation.}
\end{table}

\begin{table}
\label{tab:example}
\centering
\begin{tabular}{|c|c|c|c| }
    \hline
    \textbf{Model}  &  \textbf{R-1} & \textbf{R-2} & \textbf{R-L}\\
    \hline

    ILPSumm  &  43.37 & 16.72 & 11.26\\
    \hline
    
    ParaFuse  &  46.78 & \textbf{18.93} & \textbf{12.47}\\
    \hline
    
    Our Proposed Method  &  \textbf{47.43} & 17.28 & 10.58\\
    \hline
    
\end{tabular}

\caption{MAG-20 Extractive Evaluation.}

\end{table}

\subsection{Abstractive Evaluation}

Since metrics based on lexical overlap such as ROUGE favor extractive summaries \cite{chu2019meansum}, we conduct abstractive summary evaluation using five human evaluation metrics we propose for this study. The metrics have been developed in consultation with two co-author Linguists. The five human evaluation metrics are: 1) Entailment; 2) Coherence; 3) Conciseness; 4) Readability; and 5) Grammar. Our co-author linguists evaluated the abstractive summaries on a scale of 1 to 5 on each of the human evaluation metrics.

Our co-author linguists independently reviewed the DUC-2004 and MAG-20 results generated using our approach, ILPSumm, and ParaFuse. When determining the rating for each criterion, they used the source articles to validate the summary. Then, they used their own compiled summaries to compare to the resulting abstractive summary. The closer the abstractive summary was to the details in their notes, the higher the Entailment. The human evaluators judged Coherence by sentence structure and whether the sentences showed logical progression. When examining Conciseness, they looked for areas of the abstractive summary that were repeated. They also noted whether a sentence carried the logical progression of the paragraph. For Readability, they did not take grammar or spacing into consideration; they looked for sentence fragments, word order, and took note of instances of missing subjects or verbs. When rating Grammar, they gave the abstractive summary a lower rating for comma splices or extra spacing than if there were fragments or inappropriate punctuation.

\vspace{-0.9mm}

In addition to the human evaluation metrics, we also use copy rate\cite{nayeem2018abstractive} for evaluating abstractive summaries. Copy Rate  assesses the rate of novel word generation. As shown in Table VI, our framework achieves the lowest copy rate indicating that we are able to generate more novel words.

\begin{table}[ht]
    \centering
    \scalebox{0.8}{
    \begin{tabular}{|c|c|c|}
        \hline
        \textbf{Task} & \textbf{Model} & \textbf{Copy Rate} \\
        \hline
        \multirow{3}{*} {DUC-2004} & ILPSumm & 0.99\\
                        & ParaFuse & 0.76\\
                        & \textbf{Our Approach} & \textbf{0.68}\\
        \hline
        \multirow{3}{*} {MAG-20} & ILPSumm & 0.96\\
                        & ParaFuse & 0.88\\
                        & \textbf{Our Approach} & \textbf{0.72}\\
        \hline
    \end{tabular}}
    \caption{Copy Rate Evaluation.}
    \label{multirow_table}

\end {table}

\begin{figure}[htbp]
\centerline{\includegraphics[width=60mm]{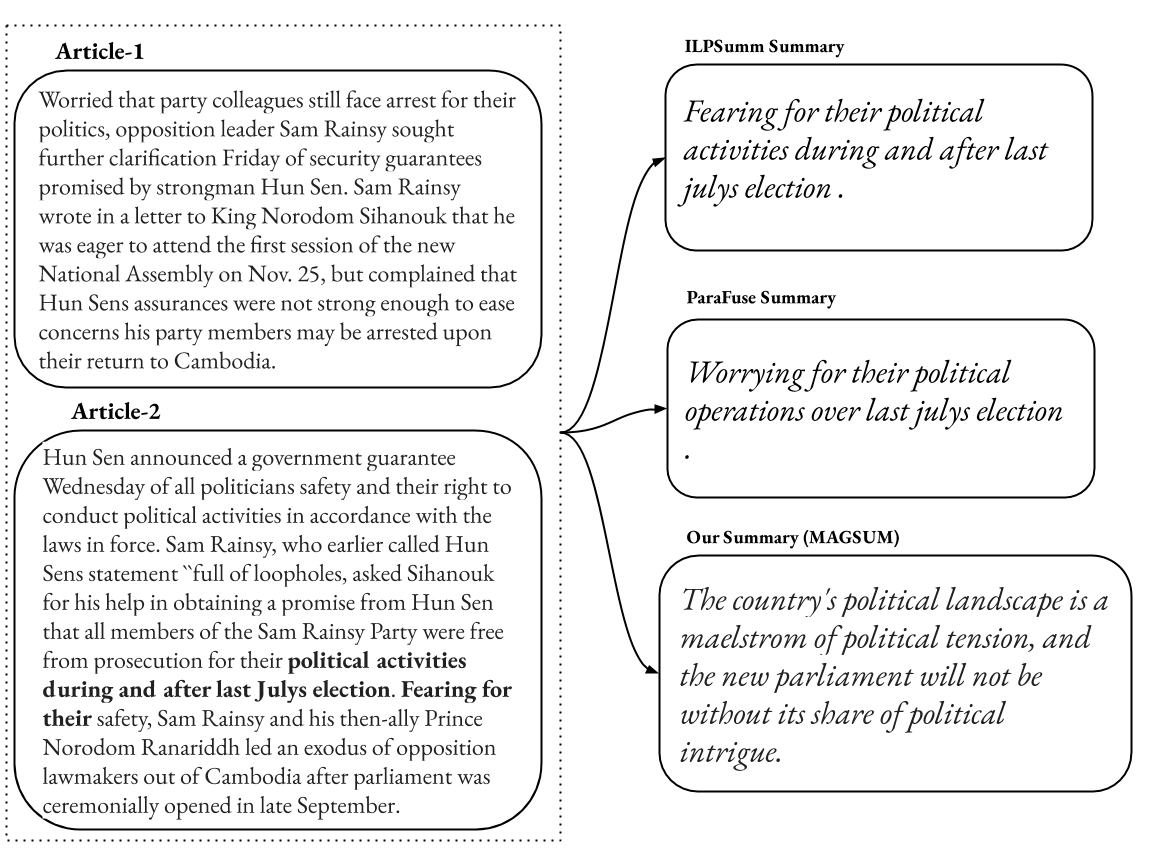}}

\caption{Comparison of abstractive summaries.}

\label{fig}
\end{figure}

\begin{table}[ht]
    \centering
    \scalebox{0.50}{
    \begin{tabular}{|c|c|c|c|c|c|c|}
        \hline
        \textbf{Human Evaluator} & \textbf{Model} & \textbf{Entailment} & \textbf{Coherence} & \textbf{Conciseness} & \textbf{Readability} & \textbf{Grammar} \\
        \hline
        \multirow{3}{*} {Evaluator-I} & ILPSumm & 0.60 & 0.26 & 0.22 & 0.20 & 0.20\\
                        & ParaFuse & 0.62 & 0.47 & 0.55 & 0.46 & 0.53\\
                        & Our Approach & \textbf{0.66} & \textbf{0.52} & \textbf{0.63} & \textbf{0.50} & \textbf{0.60}\\
        \hline
        \multirow{3}{*} {Evaluator-II} & ILPSumm & 0.50 & 0.38 & 0.34 & 0.34 & 0.40\\
                        & ParaFuse & 0.64 & 0.51 & 0.50 & 0.45 & 0.51\\
                        & Our Approach & \textbf{0.66} & \textbf{0.54} & \textbf{0.55} & \textbf{0.48} & \textbf{0.57}\\
            
        \hline
    \end{tabular}}
    \caption{DUC-2004 Abstractive Summarization Results.}
    \label{multirow_table}

\end {table}

\begin{table}[ht]
    \centering
    \scalebox{0.50}{
    \begin{tabular}{|c|c|c|c|c|c|c|}
        \hline
        \textbf{Human Evaluator} & \textbf{Model} & \textbf{Entailment} & \textbf{Coherence} & \textbf{Conciseness} & \textbf{Readability} & \textbf{Grammar} \\
        \hline
        \multirow{3}{*} {Evaluator-I} & ILPSumm & \textbf{0.89} & 0.63 & 0.71 & 0.53 & 0.38\\
                        & ParaFuse & 0.82 & 0.64 & \textbf{0.79} & 0.61 & 0.56\\
                        & Our Approach & 0.85 & \textbf{0.70} & 0.77 & \textbf{0.65} & \textbf{0.59}\\
        \hline
        \multirow{3}{*} {Evaluator-II} & ILPSumm & \textbf{0.84} & 0.71 & 0.70 & 0.65 & 0.47\\
                        & ParaFuse & 0.83 & \textbf{0.79} & 0.76 & 0.68 & 0.60\\
                        & Our Approach & 0.80 & 0.77 & \textbf{0.81} & \textbf{0.70} & \textbf{0.67}\\
            
        \hline
    \end{tabular}}
    \caption{MAG-20 Abstractive Summarization Results.}
    \label{multirow_table}

\end {table}

Experimental results show that our proposed approach performs significantly better in human abstractive evaluation metrics and copy rate. This is mainly due to the ALU generation using a fine-tuned GPT-2 model and minimizing the syntactic similarity (Equation-6) of generated ALUs.

\subsubsection{DUC-2004 Abstractive Evaluation}
For DUC-2004, our proposed approach consistently performs better than ILPSumm or ParaFuse on the 5 human evaluation criteria. ILPSumm and ParaFuse show better results in entailment. In contrast, our approach generally performs comparably across the 5 criteria. Thus, we can clearly infer generating summaries that are entailed by source articles is easier than generating summaries that are coherent, concise, readable, and grammatical. This is because if summaries have words copied from the source articles, it is highly likely that they are entailed by the source articles. Since the baseline approaches (ILPSumm, and ParaFuse) have higher copy rate, they do well in entailment. However, with our approach, having a low copy rate and generating summaries that are entailed by the sources articles is difficult; yet, our proposed approach still has the best entailment score for task DUC-2004.

\subsubsection{MAG-20 Abstractive Evaluation}
For MAG-20, our approach performs better than the baseline approaches in coherence, conciseness, readability, and grammar across two of our human evaluators, while marginally losing to the baselines according to one of our evaluators. As for entailment, ILPSumm performs the best which is attributed to the high copy rate by ILPSumm. Even though our approach generates significantly more novel words than ILPSumm or ParaFuse, we lose to the best entailment score by only 4\%.  Further, ILPSumm, ParaFuse, and our proposed approach perform generally better on MAG-20 than on DUC-2004. We surmise this is due to the headline generation task for DUC-2004, while we use author-provided titles for MAG-20.

\section{Conclusion and Future Work}
We proposed an unsupervised multi-document abstractive summarization framework that, when given a set of documents from MAG, automatically clusters the documents and then generates summaries for each cluster. Our framework consists of extractive and abstractive phases. In the extractive phase, we use coreference resolution to extract groups of inter-dependent sentences from source articles and centroid-based clustering followed by an enhanced multi-sentence compression algorithm to generate topically informative and relevant summaries. In the abstractive phase, we use text generation technique to generate abstractive language units that are synthesized into an abstractive summary. The number of summaries in our proposed method is adaptively determined based on the semantic analysis of the topics discussed in the documents. We introduce MAG-20, a dataset of topically-clustered groups of scientific articles across 20 Fields of Study and their abstractive summaries. Results show that our proposed approach performs better than state-of-the-art centroid-based summarization techniques on 5 human evaluation metrics and copy rate. In the future, we plan to use additional knowledge and metadata such as citation relationships among scientific articles for document summarization.

\section{Acknowledgment}
The authors are deeply grateful to Daniel Foose for helping with developing scripts for efficient data collection from MAG.

\bibliographystyle{IEEEtran}  
\bibliography{bibliography.bib}

\end{document}